\pdfoutput=1

\documentclass[11pt]{article}

\usepackage[]{emnlp2023}

\usepackage{times}
\usepackage{latexsym}

\usepackage{xcolor}         
\usepackage{colortbl}         
\usepackage{amssymb} 

\usepackage{booktabs}

\usepackage[T1]{fontenc}

\usepackage[utf8]{inputenc}

\usepackage{svg}

\usepackage{microtype}

\usepackage{graphicx}

\title{In-Context Learning for Text Classification with Many Labels}

\author{Aristides Milios$^1$, Siva Reddy$^{1,2,3}$, Dzmitry Bahdanau$^{1,2}$\\
Mila and McGill University$^1$, ServiceNOW Research$^2$, Facebook CIFAR AI Chair$^3$\\
\texttt{\{aristides.milios,siva.reddy,bahdanau\}@mila.quebec}}

\begin{document}
\maketitle
\begin{abstract}
In-context learning (ICL) using large language models for tasks with many labels is challenging due to the limited context window, which makes it difficult to fit a sufficient number of examples in the prompt. In this paper, we use a pre-trained dense retrieval model to bypass this limitation, giving the model only a partial view of the full label space for each inference call. Testing with recent open-source LLMs (OPT, LLaMA), we set new state of the art performance in few-shot settings for three common intent classification datasets, with no fine-tuning. We also surpass fine-tuned performance on fine-grained sentiment classification in certain cases. 
We analyze the performance across number of in-context examples and different model scales, showing that larger models are necessary to effectively and consistently make use of larger context lengths for ICL. 
By running several ablations, we analyze the model's use of: a) the similarity of the in-context examples to the current input, b) the semantic content of the class names, and c) the correct correspondence between examples and labels. We demonstrate that all three are needed to varying degrees depending on the domain, contrary to certain recent works.
\end{abstract}

\section{Introduction}

In-context learning (ICL) using large language models (LLMs) has recently exploded in popularity. Models pre-trained on massive amounts of textual data are able to reach reasonable performance on a wide variety of tasks with only a few examples of input and output for a given task provided in the model's input prompt in natural language \cite{gpt3,gopher,palm}. In this work, we study whether ICL can handle challenging classification tasks with many possible labels, by augmenting the LM with a secondary pre-trained retrieval model.

\begin{figure*}[h]
    \centering
    \includegraphics[scale=0.6]{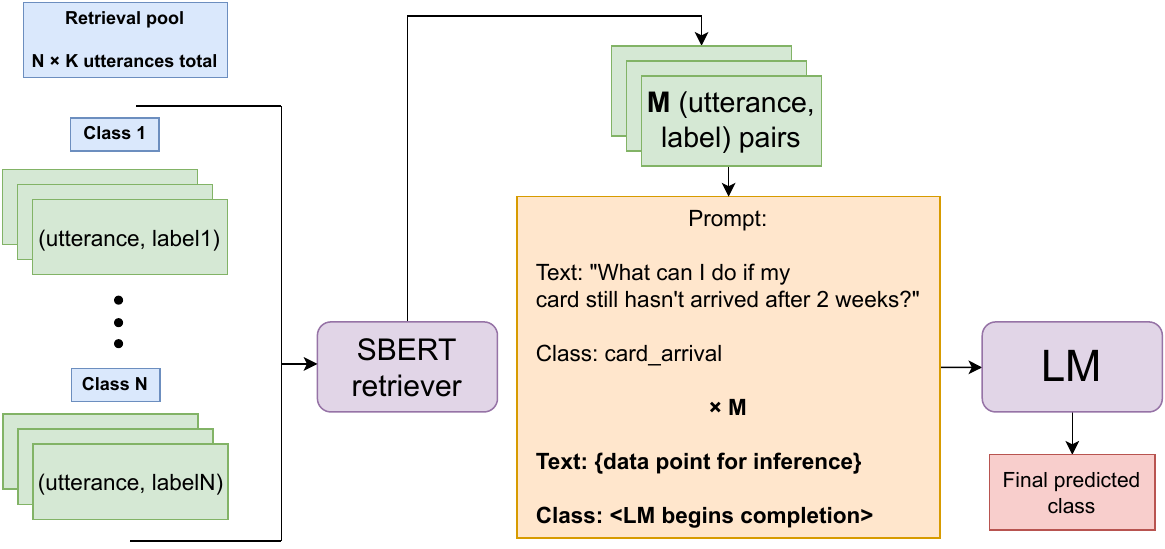}
    \caption{Complete pipeline for intent detection with retrieval-augmented in-context learning}
    \label{fig:pipe}
\end{figure*}

The main problem with applying ICL to tasks involving classification with many labels is the limited context window these models have. Ordinarily with ICL, at minimum one example from each class is provided in-context to allow the model to make a choice between all the labels of the task. Because of this limitation, ICL has not been directly applied to these sorts of problems. In this work we relax this requirement, allowing the model to see only a subset of the most relevant labels for the given datapoint we are performing inference on.  By testing on intent classification (upwards of 50 classes) and fine-grained sentiment analysis (upwards of 25 classes), we demonstrate that the resulting performance with this method can reach SoTA. By coupling the LLM with an external pre-trained dense retriever model \cite{sbert,dpr}, we can dynamically retrieve a set of examples to provide to the LM in-context, that reflects only the most relevant labels to the current example in the label space. Most existing work on augmenting LMs with retrieval models \cite{ram2023incontext, shi2023replug} focuses on tuning the retrieval and/or LM. We demonstrate that even without tuning either, when the pre-trained models are strong enough we can still achieve SoTA across various tasks using ICL.

We evaluate LLMs in this setting with three intent classification datasets: BANKING77 \cite{bankingdataset}, HWU64 \cite{hwudataset}, and CLINC150 \cite{clincdataset}, as well as one fine-grained sentiment classification dataset: GoEmotions \cite{goemotions}. Experiments are done using the LLaMA models \cite{touvron2023llama} and the OPT models \cite{zhang2022opt} as LLMs. We compare the performance achieved against adapter-based fine-tuning of MLM models (DeBERTa-v2-XXLarge with the ``Pfeiffer'' bottleneck-style adapter \cite{pfeiffer-etal-2020-mad} implemented with AdapterHub \cite{pfeiffer2020AdapterHub}) and the previous SoTA for intent detection (ConvFit; \citealt{vulic-etal-2021-convfit}), as well as comparing against SetFit \cite{setfit}, a recent lightweight method involving contrastive training of small MLM models.

The contributions of this work are:

\begin{enumerate}
    \item We show that retrieval-augmented ICL is an effective way to tackle text classification tasks with many labels without additional tuning of either the retriever or the LM, either matching or outperforming fine-tuned adapter-based and contrastive-pre-training-based methods. Notably, truncating the dataset by showing only a subset to the LM at a time does not prevent us from achieving SoTA performance, and allows us to apply LLMs to problems that they have not been applied to before,
    \item We analyze ICL performance over different numbers of examples and demonstrate that larger models better are able to take advantage of more examples in-context than smaller models, which mostly plateau and/or see decreasing performance,
    \item We perform several ablation studies to determine what aspects of the inputs and outputs the model is using for ICL. Certain recent works investigating ICL \cite{rethinking-role-demonstrations, razeghi2022impact} have recently called into question how much models are actually ``learning'' with ICL and what they are learning from. We ablate three different elements (semantic label names, correct input-output correspondences, and semantically similar demonstrations to the current input). Contrary to this emerging literature, our experiments demonstrate that they are all used to varying degrees, depending on the dataset and domain.
\end{enumerate}

\section{Method}

\paragraph{Retrieval-Augmented ICL:} Our setup assumes $N$ classes (unique labels) with $K$ examples in each class. Each example is composed of an \texttt{(input, label)} tuple. We assume that we have a limited number of examples $M$ to fit in the prompt, based on the model's context length. $M$ can be fixed or based on ``saturating'' the prompt greedily by selecting examples until we run out of room in the context window. From our total pool of examples of size $N \times K$, we retrieve the $M$ examples using the cosine similarity values given by our retrieval model. Having retrieved our $M$ examples, we then produce the prompt by concatenating the \texttt{(input, label)} tuples in a set prompt format (see Figure \ref{fig:pipe}), similar to existing in-context learning setups. The final prediction is then taken from the LM by having it produce a continuation based on our prompt. A full visual description of the retrieval process is visible in Figure \ref{fig:pipe}.

\paragraph{Retrieval model:} The retrieval model used is a Sentence-BERT model trained in a Siamese dual-network setup to be able to retrieve text based on cosine similarity of the embedding vectors it produces, described in \citet{reimers-gurevych-2019-sentence}. The model we use is a contrastively trained model which has been pre-trained on a massive generic dataset of text pairs. We use the retrieval model as-is in all experiments.  Cosine similarity is used to retrieve examples from the retrieval pool of examples (tested in 5-shot and 10-shot scenarios, signifying the number of examples from each class in the retrieval pool).

\section{Experimental Setup}

\paragraph{Specific retrieval model:} For our sentence encoder/retriever, we use the SentenceTransformers library \cite{sbert}, and use the pre-trained ``all-mpnet-base-v2'' model (a 110M parameter model pre-trained on over 1 billion training pairs). The SetFit results are based on contrastively tuning the same pre-trained model trained by Microsoft through the Setfit library\footnote{https://github.com/huggingface/setfit}.

\paragraph{Prompt saturation:} The number of examples that fit in-context when greedily filling the context window depends on the specific dataset. For the intent detection datasets, this number was around 110 examples. For GoEmotions, this number was around 70 (140 using the full 4K context length of the LLaMA-2 models).

\paragraph{Splits:} For the intent detection experiments, to allow for direct comparison with previous works, we use the same 5-shot and 10-shot sets as DialoGLUE \cite{dialoglue}. Experiments are run 3 times and the accuracies are averaged, except the zero-training LLM setups, which are deterministic. For the GoEmotions experiments we average the results across 3 different random 10 and 5-shot splits, as no pre-existing few-shot splits exist. The GoEmotions experiments are composed of the subset of GoEmotions data (84\% of training set, 85\% of testing set) where the there is only one emotion label, to avoid issues of enforcing an ordering on a linearized version of multiple labels in sequence, as well as to mimic the single-label intent detection datasets setup more closely. Default library parameters were used. 

\paragraph{Computing Hardware and model differences:} All experiments were performed on a single A100 80GB GPU, except those with OPT 175B, which were performed with 8 A100 GPUs. For LLaMA 65B and 70B 8-bit quantization was used. The main difference between the OPT and LLaMA models is the amount of pre-training data used. The LLaMA models were trained on 1T-1.4T tokens, while the OPT models were only trained on 180B tokens (see \cite{zhang2022opt} and \cite{touvron2023llama} for more details). LLaMA-2 models were trained on 2T tokens.

\renewcommand{\arraystretch}{1.2}

\begin{table*}[h]
  \centering
  \small
  \caption{Intent classification accuracy for retrieval+ICL and baseline methods. All retrieval+ICL results are with 20 in-prompt examples unless otherwise specified. The retrieval/training dataset size is given by the second row of the header (10-shot is 10 examples per class, 5-shot is 5).}
    \begin{tabular}{lcccccc}
    \toprule
    \textbf{Model}
          & \multicolumn{2}{c}{\textbf{BANKING 77}} & \multicolumn{2}{c}{\textbf{HWU 64}} & \multicolumn{2}{c}{\textbf{CLINC 150}} \\
    \cmidrule(lr){2-3} \cmidrule(lr){4-5} \cmidrule(lr){6-7}
          & \textbf{5-shot} & \textbf{10-shot} & \textbf{5-shot} & \textbf{10-shot} & \textbf{5-shot} & \textbf{10-shot} \\
    \midrule
        Pre-trained SBERT 1-NN & 78.41 & 85.39 & 69.89 & 75.46 & 82.51 & 84.84 \\
        ConvFit (reported) & - & 87.38 & - & 85.32 & - & 92.89 \\
        SetFit & 79.89 {\scriptsize $\pm$ 0.14} & 84.51 {\scriptsize $\pm$ 0.60} & 78.38 {\scriptsize $\pm$ 0.73} & 83.35 {\scriptsize $\pm$ 0.57} & 88.68 {\scriptsize $\pm$ 0.20} & 90.67 {\scriptsize $\pm$ 0.29} \\
        DeBERTa (Pfeiffer) & 81.47 {\scriptsize $\pm$ 1.6} & 88.41 {\scriptsize $\pm$ 0.19} & 79.80 {\scriptsize $\pm$ 0.81} & 86.93 {\scriptsize $\pm$ 0.052} & 91.86 {\scriptsize $\pm$ 0.66} & 95.05 {\scriptsize $\pm$ 0.33} \\ \hline
        OPT 13B & 81.23 & 85.65 & 78.90 & 83.64 & 85.27 & 89.24 \\
        OPT 175B & 81.30 & 86.14 & 83.74 & 84.94 & 90.96 & 93.09 \\
        LLaMA 7B & 84.42 & 87.63 & 85.87 & 87.55 & 88.58 & 91.73 \\
        LLaMA 65B & 87.73 & 90.71 & 89.03 & 90.06 & 91.89 & 94.47 \\ \hline
        LLaMA 2 7B & 86.40 & 89.45 &  87.55 & 87.82  & 94.13 & 95.20 \\
        LLaMA 2 7B 4K & 85.91 & 89.48 & 87.17 & 90.33 & 95.35 & 96.02 \\
        LLaMA 2 70B & 87.56 & 90.58 & 88.20 & 89.77 & 96.42 & 97.13 \\
        LLaMA 2 70B 4K & \textbf{88.96} & \textbf{92.11} & \textbf{90.61} & \textbf{91.73} & \textbf{97.56} & \textbf{98.18} \\
    \bottomrule
    \end{tabular}%
    \label{tab:allperf}
\end{table*}%

\begin{table*}[h]
  \centering
  \small
  \caption{Sentiment classification macro F1 score (following prior work) over 3 random splits for retrieval+ICL and baseline methods. All retrieval+ICL results are from saturating the prompt with in-prompt examples (with a 2K prompt length unless otherwise specified). The retrieval/training dataset size is given by the second row of the header (10-shot is 10 examples per class, 5-shot is 5). $+$Neut refers to the case where the ``neutral'' class (lack of emotion) is included in the dataset. }
    \begin{tabular}{lcccc}
    \toprule
    \textbf{Model}
          & \multicolumn{4}{c}{\textbf{GoEmotions}} \\
    \cmidrule(lr){2-5}
          & \textbf{5-shot} & \textbf{10-shot} & \textbf{5-shot $+$Neut} & \textbf{10-shot $+$Neut} \\
    \midrule
        Pre-trained SBERT 1-NN & 9.48 {\scriptsize $\pm$ 0.58} & 11.02 {\scriptsize $\pm$ 1.0} & 7.55 {\scriptsize $\pm$ 0.79} & 8.38 {\scriptsize $\pm$ 0.48} \\
        SetFit & 25.44 {\scriptsize $\pm$ 4.5} & 34.69 {\scriptsize $\pm$ 3.6} & 21.40 {\scriptsize $\pm$ 3.18} & 27.78 {\scriptsize $\pm$ 0.73} \\
        DeBERTa (Pfeiffer) & 18.43 {\scriptsize $\pm$ 2.9} & 32.33 {\scriptsize $\pm$ 0.77} & 13.86 {\scriptsize $\pm$ 1.49} & 25.42 {\scriptsize $\pm$ 1.9} \\ \hline
        LLaMA 7B & - & - & 22.99 {\scriptsize $\pm$ 0.64} & 24.61 {\scriptsize $\pm$ 0.47} \\
        LLaMA 65B & - & - & 24.31 {\scriptsize $\pm$ 0.73} & 25.63 {\scriptsize $\pm$ 0.86} \\ \hline
        LLaMA 2 7B & 29.60 {\scriptsize $\pm$ 1.5} & 31.40 {\scriptsize $\pm$ 0.83} & 23.78  {\scriptsize $\pm$ 1.1} & 24.75 {\scriptsize $\pm$ 0.43} \\
        LLaMA 2 7B 4K & 28.01 {\scriptsize $\pm$ 1.2} & 30.33 {\scriptsize $\pm$ 1.64}  & 23.79 {\scriptsize $\pm$ 1.9} & 23.57 {\scriptsize $\pm$ 0.52} \\
        LLaMA 2 70B & \textbf{36.14} {\scriptsize $\pm$ 1.7} & \textbf{37.81} {\scriptsize $\pm$ 1.3} & 24.20 {\scriptsize $\pm$ 0.13} & 25.29 {\scriptsize $\pm$ 0.42} \\
        LLaMA 2 70B 4K & - & 37.17 {\scriptsize $\pm$ 0.37} & \textbf{28.26} {\scriptsize $\pm$ 0.19} & \textbf{29.10} {\scriptsize $\pm$ 0.68} \\
        LLaMA 2 70B 4K Retrieval w/o Neutral & - & - & - & 28.95 {\scriptsize $\pm$ 0.52} \\
    \bottomrule
\end{tabular}
    \label{tab:perfgoem}
\end{table*}%

\paragraph{Restricting model output:} To reduce computational load and make inference easier, instead of using the logits of the LLM to rank our many classes (requiring multiple forward passes, as class names consist of multiple tokens), we let the LLM generate freely. Having generated an output text, we then use the retrieval model (SBERT) to retrieve the most similar class label from our set of classes. This allows us to restrict the model output to the set of classes we want without incurring additional inference cost. Instances of generated predictions that do not match our class list are few regardless, and shrink proportionately to the number of examples provided in-context.

\paragraph{Baselines:} Several baselines are provided. The baseline ``Pre-trained SBERT 1-NN'' refers to using the SBERT retrieval model to retrieve the most similar example in the retrieval pool and use its label directly as the prediction (1-nearest-neighbor). The ConvFit baseline is taken from the reported numbers in the ConvFit paper directly. The baseline ``DeBERTa (Pfeiffer)`` is the DeBERTa-XXL model released by Microsoft, trained via AdapterHub with the Pfeiffer-style bottleneck adapters \cite{pfeiffer-etal-2020-mad, pfeiffer2020AdapterHub}. Preliminary results with  other adapter types (LoRA, IA\textsuperscript{3}, etc.) showed that the Pfeiffer-style adapters were the most effective in this particular use-case. The DeBERTa-XXL model was fine-tuned until performance saturation (early stopping). SetFit \cite{setfit} results are also provided, a method involving contrastive fine-tuning of a retriever model with a classification head, as it is also a competitive and lightweight baseline in this setup. The selection of baselines was done based on recent strong progress on few-shot classification using parameter-efficient fine-tuning, in certain cases having been shown to perform better than full fine-tuning \cite{fewshotadapters}.

\section{Results}

\paragraph{Example ordering:} We provide a brief study regarding how to order examples in-prompt by similarity, since previous work has been inconclusive on this front, suggesting that the ideal ordering is dataset dependent \cite{liu-etal-2022-makes}. As seen from Table \ref{tab:ordering}, least-to-most (LTM) similar was the most effective ordering across all datasets. Larger models are significantly less sensitive to ordering.

\begin{table*}[]
  \centering
  \small
  \caption{Comparison of LLaMA 7B and OPT 13B model prompt orderings on intent detection datasets (20 examples in prompt, 10-shot), random split. MTL is most-to-least similar and LTM is the inverse.}
    \begin{tabular}{lccccccccc}
    \toprule
    \textbf{Model}
          & \multicolumn{2}{c}{\textbf{BANKING}} & \multicolumn{2}{c}{\textbf{HWU}} & \multicolumn{2}{c}{\textbf{CLINC}} & \multicolumn{3}{c}{\textbf{GoEmotions}}\\
    \cmidrule(lr){2-3} \cmidrule(lr){4-5} \cmidrule(lr){6-7} \cmidrule(lr){8-10}
          & \textbf{MTL} & \textbf{LTM} & \textbf{MTL} & \textbf{LTM} & \textbf{MTL} & \textbf{LTM} & \textbf{MTL} & \textbf{Random} & \textbf{LTM} \\
    \midrule
    OPT 13B & 73.64 & \textbf{85.65} & 76.39 & \textbf{83.64} & 81.11 & \textbf{89.24} & - & - & - \\
    LLaMA 7B & 83.64 & \textbf{87.63} & 86.99 & \textbf{87.55} & 90.20 & \textbf{91.73} & 15.91 & 20.89 {\scriptsize $\pm$ 0.85} & \textbf{23.58} \\
    LLaMA 65B & 88.08 & \textbf{90.71} & 89.03 & \textbf{90.06} & 93.47 & \textbf{94.47} & - & - & - \\
    \bottomrule
    \end{tabular}%
  \label{tab:ordering}%
\end{table*}%


\paragraph{SoTA performance:} Tables \ref{tab:allperf} and \ref{tab:perfgoem} shows the performance comparison of all methods. Performance of the retrieval+ICL pipeline on BANKING, HWU and CLINC is state of the art in both the 5 and 10-shot settings. Not only this, but to significantly surpass the previous state of the art for all three intent classification datasets only LLaMA-2 7B is necessary, which with 8-bit quantization can be run on consumer hardware. In the most challenging evaluation setting (the highly-specialized intent classes of the BANKING dataset in the most data-scarce 5-shot setting), the margin between DeBERTa and LLaMA-2 70B is 7.49\%. In general the DeBERTa model showed lower performance in the 5-shot scenarios, likely due to the extremely limited data. In the case of GoEmotions (Table \ref{tab:perfgoem}), when using the neutral category, the Retrieval+ICL pipeline manages to clearly win against the strongest baseline (SetFit) only in the 5-shot case. In the 10-shot case, we can see that Retrieval+ICL performs at least on par, but more likely better than SetFit. Table \ref{tab:example_io} shows the difficulty of the GoEmotions task, specifically with regards to how granular the classes are.

\begin{figure}
    \centering
    \includegraphics[scale=0.4]{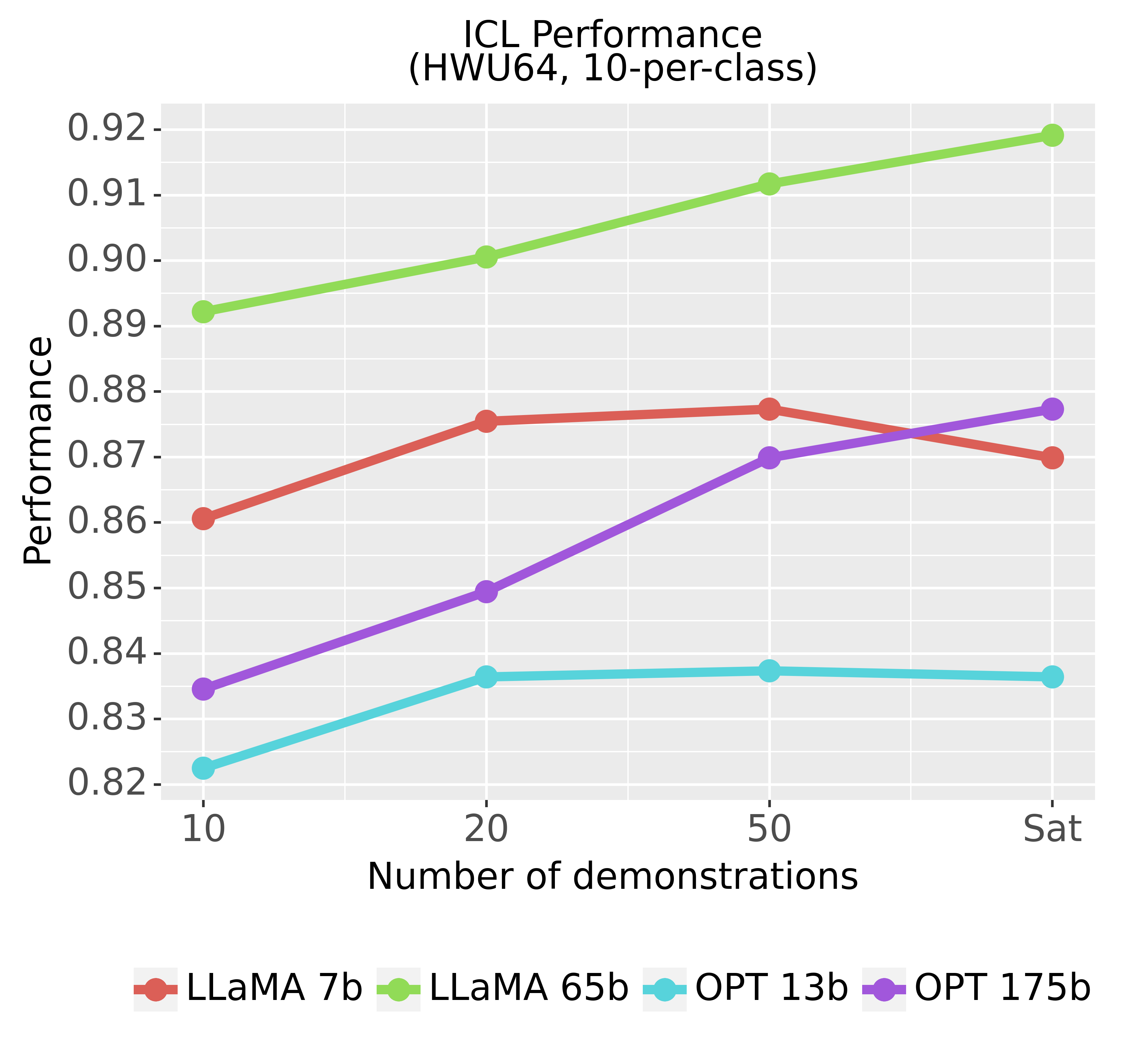}
    \caption{HWU performance as a function of the number of examples in prompt. The x-axis scale is non-linear, meaning that there are diminishing returns with more examples. ``Sat'' (saturated) indicates filling the prompt greedily until the max length is reached.}
    \label{fig:hwu_exs}
\end{figure}

\paragraph{Performance degredation:} We also provide a study of how performance changes given the number of examples provided in-context. Figure \ref{fig:hwu_exs} shows this variation for the HWU64 dataset. The x-axis value of 110 indicates a fully saturated context window, which is on average this number of examples. In the case of LLaMA-7B, performance somewhat degrades after a certain number of demonstrations. Looking at Tables \ref{tab:allperf} and \ref{tab:perfgoem}, comparing LLaMA-2-7B and LLaMA-2-70B in the regular and 4K context window scenarios, we see very clearly that only the 70B model is able to continually improve with the full 4K context. The 7B model instead sees matching (no improvement) or degraded performance in most cases.

\paragraph{Impact of ``Neutral'' on GoEmotions:} From the results in Table \ref{tab:perfgoem}, by comparing the results with and without the ``neutral'' category, we see that the difference between the baselines and Retrieval+ICL grows, implying that ``neutral'' disproportionately hurts the Retrieval+ICL performance. We note that correctly predicting the neural class is challenging for the LM. We demonstrate that removing ``neutral'' from the retrieval pool does not harm performance (``Retrieval without Neutral'' in Table \ref{tab:perfgoem}). Analyzing the results for one of the runs, we see that out of the 1605 examples of the ``neutral'' class in the test set, ``neutral'' only appears in the top 3 classes retrieved by the retriever (by number of examples) only 9\% of the time (in the top 5 classes 18\%). This suggests that the retriever may be limiting the performance.

\section{Ablation Studies}

\begin{figure}
    \centering
    \includegraphics[scale=0.53]{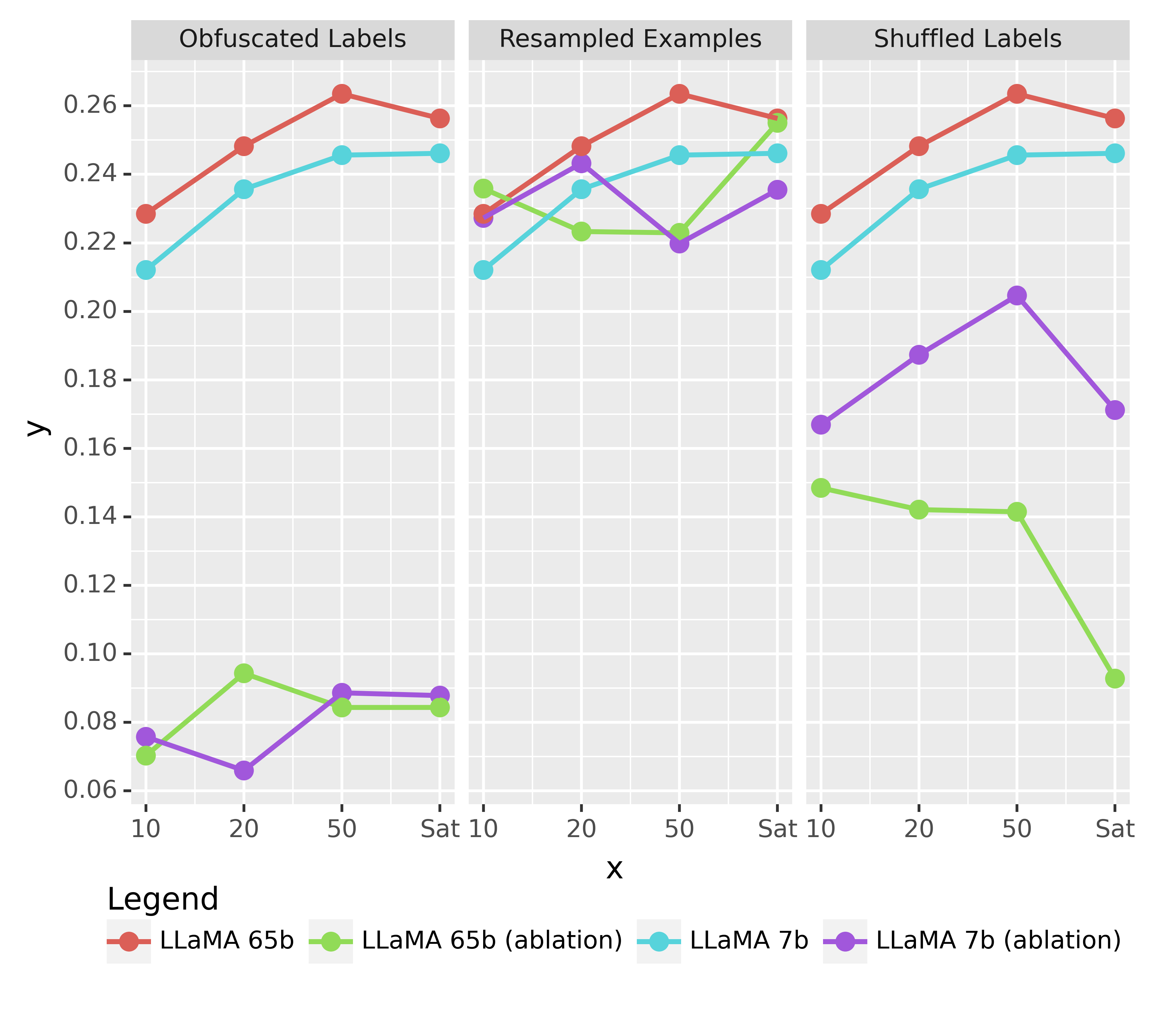}
    \caption{Classification accuracy for three ablations for GoEmotions: obfuscated labels (left), resampled in-context examples (center), shuffled labels (right).}
    \label{fig:resampling_goem} \label{fig:obfuscating_goem} \label{fig:shuffling_goem}
\end{figure}

Several ablations studies are done to test what aspects of the retrieved examples the LLM is using to make the predictions. The ablation studies were done on a random split of the HWU dataset and the GoEmotions dataset. Ablation results for HWU are shown visually in Figure \ref{fig:resampling} and for GoEmotions in Figure \ref{fig:resampling_goem}.

\begin{enumerate}
    \item \textbf{Obfuscated labels:} We change all the class names to randomly set enumerated names (``Class 1'', ``Class 2'', etc.). The intent is to disentangle the model's use of prior (pre-training) knowledge to perform the task (based on the semantic content of the label names) from the input-output provided in the prompt.
    \item \textbf{Resampled in-context examples:} To test if similarity between the demonstrations provided in the prompt and the current input example is actually necessary for effective performance. By resampling from the classes initially retrieved by the retriever model, we preserve the distribution of labels but change the input demonstrations themselves so that they are no longer the nearest in the embedding space for each class.
    \item \textbf{Shuffled labels:} Similarly to \citet{rethinking-role-demonstrations}, after the retrieval step we shuffle the correspondence between the inputs and labels of the retrieved examples, such that inputs are matched randomly from the set of labels the inputs originally belonged to. The intent of this ablation is to examine if the model requires correct input-label correspondences (something that \citet{rethinking-role-demonstrations} calls into question), or if the model is simply using structural (e.g. prompt format) and distributional (e.g. the distribution of labels in the prompt) elements to produce a prediction.
\end{enumerate}

\section{Discussion}

\subsection{Small models cannot use long contexts as effectively as large models}

One trend noticeable from the performance graph as a function of the number of examples for HWU (see Figure \ref{fig:hwu_exs}) is that small models seem to be unable to use more examples as effectively as large models. The smaller OPT model is unable to effectively make use of the entire context window when it is filled and remains at relatively low performance. In contrast, OPT 175B shows continual improvement when more examples are added. A similar trend is visible for the LLaMA models, where the performance of the 7B model does not change significantly (see \ref{fig:hwu_exs}), but the 65B model is able to continuously improve. The smaller models either level off (OPT-13B) or lose performance (LLaMA-7B). In the 4K full context window settings for LLaMA-2, the difference between model scales is even more apparent (Tables \ref{tab:allperf} and \ref{tab:perfgoem}). We see the small model showing inconsistent use of the longer contexts; sometimes improving, but mostly staying the same or worsening performance. Meanwhile, the large model consistently improves with the full context in almost all cases.

\begin{figure}
    \centering
    \includegraphics[scale=0.53]{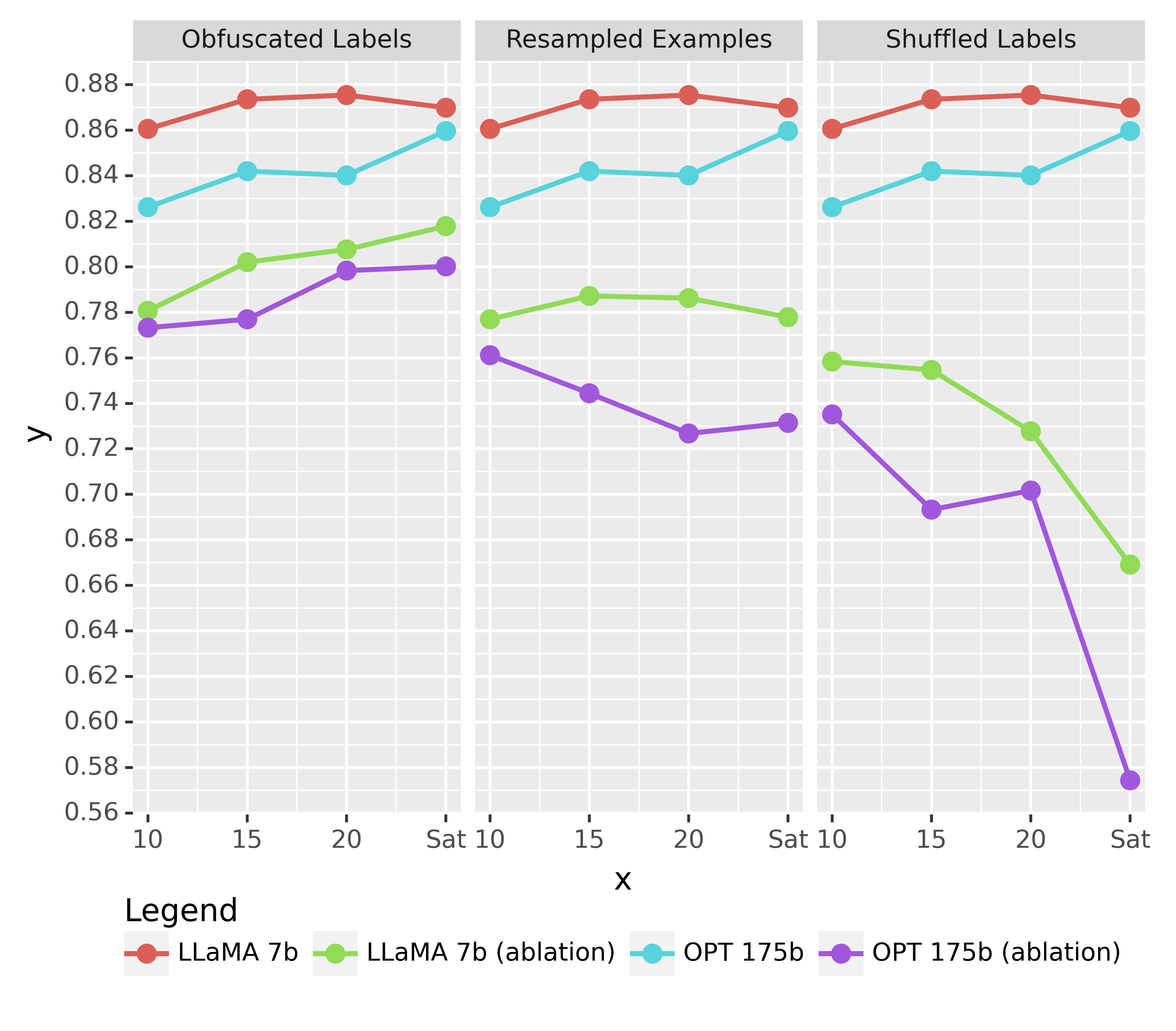}
    \caption{Classification accuracy for three ablations for HWU64: obfuscated labels (left), resampled in-context examples (center), shuffled labels (right).}
    \label{fig:resampling} \label{fig:obfuscating} \label{fig:shuffling}
\end{figure}

\subsection{Similarity to current datapoint matters for intent classification}

In the resampling ablation for HWU (see Figure \ref{fig:resampling}) we see that resampling from the initial class distribution provided by the retriever model damages the performance across both OPT 175B and LLaMA 7B. This supports the strong performance numbers of the LLMs, showing that the similarity between in-context demonstrations and the current input matters. This implies that the LM is doing more than just selecting the most common class or just using the shortlist of class labels from the full set of classes to select in a more zero-shot fashion. One interesting difference to note is that OPT 175B, the larger model, shows a larger drop from the resampling as the number of in-context demonstrations increases, compared to LLaMA-7B, whose performance stays roughly constant (but lower than non-resampled). This may indicate that the LLaMA models with their additional training data are more robust to the resampling process, due to stronger pre-training knowledge and/or more robust performance overall. In the case of GoEmotions, we see almost no variation with resampling, showing that similarity to the input example is less influential, though the ordering of the examples relative to each other does seem to make a difference for the 7B model (Table \ref{tab:ordering}).

\begin{table*}[]
  \centering
  \small
  \caption{Sample datapoints from GoEmotions}
    \begin{tabular}{p{10cm}|p{2cm}|p{2cm}}
    \toprule
    \textbf{Text} & \textbf{Prediction LLaMA-2-70B} & \textbf{Gold label} \\
    \midrule
    Lmao the brigading is real                                                                                                                        & amusement              & amusement    \\
    Enjoy the void                                                                                                                                    & neutral                & neutral      \\
    I really relate to this.                                                                                                                          & realization            & approval     \\
    This is the only viable way out of Brexit.                                                                                                        & optimism               & approval     \\
    want* a source on that, sorry.                                                                                                                    & desire                 & remorse      \\
    I didn't know that, thank you for teaching me something today!                                                                                    & gratitude              & gratitude    \\
    Well it obviously helps you rationalize your total unwillingness to take 
    action to make the world a better place. 
    I hope that you grow past that. & sadness                & admiration   \\
    Damn, we need healthy PGs.                                                                                                                        & sadness                & annoyance    \\
    Welcome to The Church of Jesus Christ of Latter Day Saints, 
    where families can be SEPARATED forever                                               & sadness                & gratitude   \\  
    \bottomrule
    \end{tabular}%
  \label{tab:example_io}%
\end{table*}%

\subsection{Semantically significant label names matter greatly for sentiment classification}

In the obfuscation ablation (see Figure \ref{fig:obfuscating}), we see that all models are hurt by obfuscating label names. We see however that models are still able to learn to perform the task effectively, and in fact show similar improvement curves with increasing number of examples, just with a lower starting performance. This demonstrates that the semantic content of the labels is significantly useful to the models but simultaneously it is not integral to performing the task, which can also be done without semantically significant labels. In the case of GoEmotions, we see that the obfuscated labels particularly hurt the model, bringing it down significantly.It seems to be the case that the class names are integral to performance, but at the same time more examples are still helpful to the model, as in the 4K context window it still sees improved performance.

\subsection{Input-label correspondence matters for all datasets}

Shuffling the input-label correspondence is the ablation in which we see the performance of all the models decrease the most in the intent detection case (see Figure \ref{fig:shuffling}). Specifically, we see that the performance drop is proportional to the number of examples (more shuffled examples brings a larger drop). That being said, it is noteworthy that the performance of both models in this shuffled regime is still significantly above random chance for every number of demonstrations shown, implying perhaps that the LM's prior knowledge based on the label names is still contributing significantly to performance. In all 4 datasets (intent classification and GoEmotions), shuffling the labels hurts the large model more in particular. This aligns with the results of \citet{wei2023larger}, whose authors show that larger models are more able to learn perturbed input correspondences than smaller models, which manifests in this experiment as lower performance. In other words, the larger model is trying to learn the perturbed input correspondence, and thus losing more and more performance with more examples, while the smaller model is able to more effectively ignore the perturbation.

\section{Retriever and LM Generalization}

One interesting result from our experiments is the fact that generic retrievers seem to be able to quite effectively generalize across domains and tasks. Using the same exact retriever model across 3 different intent detection datasets (which according to the taxonomy of \citet{hupkes2023stateoftheart} constitutes cross-task generalization) as well as a sentiment classification dataset (according to the previous taxonomy, a cross-domain generalization) demonstrates SoTA or better performance in almost all cases. The distribution shift locus,  for both the retriever and the language model generating the final prediction, is from pretraining to testing time. This is because they are both pre-trained on massive generic data before being tested in a zero-shot setting.

\section{Related Work}

\paragraph{Nearest neighbor selection of in-context examples:} One of the earliest studies of the role of example selection in ICL is ``KATE'' \cite{liu-etal-2022-makes}. In this paper, the authors probe the performance of GPT-3 on NLP tasks using KNN retrieval (RoBERTa) for example selection. They compare this method against random selection and using the retrieval model directly (plain KNN). They also examine the effect of example ordering on performance and conclude that the most performant ordering (least-to-most and most-to-least similar orderings are tested) depends on the dataset. In our work, we also experiment with example ordering, and conclude that least-to-most ordering is the most effective across all datasets tested.

\paragraph{Works demonstrating order instability:} Several recent works have demonstrated that the order of in-context examples makes a larger difference in performance, including \citet{lu-etal-2022-fantastically, zhao2021calibrate}. These works demonstrate such order instability that certain permutations bring near SoTA performance on tasks while others perform at near random guessing.

\paragraph{Fine-tuned retrieval:} Several works employ the use of fine-tuned retrievers, re-rankers, and/or LMs, including \citet{rubin-etal-2022-learning, ram2023incontext, shi2023replug}. Some, like REPLUG \cite{shi2023replug}, use LM feedback in the form of using the LM to score documents to train the retriever. The goal of both \citet{ram2023incontext} and \citet{shi2023replug} is to improve language modeling and not ICL ability. \citet{rubin-etal-2022-learning} uses a similar LM-score-based feedback to train a retriever (like REPLUG) but for ICL. The difference between all of these works and this work is that we demonstrate that an off-the-shelf retriever is sufficient out-of-the-box for SoTA performance with no additional tuning.

\paragraph{Works calling into question efficacy of ICL:} Certain recent works have called into question the efficacy of ICL and models' ability to learn tasks they were not exposed to during pre-training \cite{rethinking-role-demonstrations, razeghi2022impact}. In \citet{rethinking-role-demonstrations} authors show that randomly perturbing input-label pairings for some tasks can still lead to reasonably good performance, calling into question whether any ``learning'' is happening at all with ICL. The work in \citet{razeghi2022impact} demonstrates that models perform better on data instances they have seen frequently during pre-training, implying that models are primarily memorizing and that their generalization capabilities in terms of ICL remain limited. \citet{xie2022explanation} suggests that ICL ability emerges due to the specific structure of the training data, specifically long-range dependencies.

\paragraph{Use of long contexts:} Several works have demonstrated that long contexts are difficult for LMs to handle and show certain peculiarities. \citet{kazemnejad2023impact} investigates the relationship between length generalization and positional embedding types, showing that in certain cases no positional embeddings can perform better. This work is closely related to use of long contexts for ICL, as it demonstrates the difficulty involved in generalizing to long context lengths, as well as providing an explanation for LMs' sensitivity to ordering (positional embeddings). In \citet{liu2023lost}, the authors investigate the impact of long contexts on document question answering, finding that the positions of the answers within the context matter greatly for performance, and generally demonstrating that longer contexts cause lower performance. In this work we show that larger models are needed to effectively take advantage of long contexts for ICL.

\paragraph{Few-shot intent detection:} The current state of the art in few-shot intent detection is the ConvFit method \cite{vulic-etal-2021-convfit}. ConvFit uses a pre-trained LM in a dual-encoder configuration (e.g. BERT or RoBERTa) with two training stages. The first stage is a conversational fine-tuning stage using a generic conversational corpus with a retrieval task (using tuples of (context, response) retrieve the correct response for each context). The second stage is fine-tuning on the specific intent classification dataset with a contrastive loss, allowing the resulting LM to be used in a KNN fashion.

\section{Conclusion}

In this work, we show that ICL with off-the-shelf frozen pre-trained retriever models can provide strong performance for text classification tasks with many labels. We show state of the art performance across three different intent classification datasets, and competitive performance with fine-grained sentiment classification. We also show that larger models are necessary to make use of more in-context examples, whereas small models mostly plateau or even show decreasing performance after a point. Through several ablation experiments, we demonstrate that LMs make use of all aspects of the input examples: semantically significant label names, correct input-label correspondences, as well as the similarity between the in-context demonstrations and the current input point, however to varying degrees depending on the dataset and domain.


 \section{Acknowledgement}

SR is supported by the Canada CIFAR AI Chairs program and the NSERC Discovery Grant program. AM is supported by an IVADO Excellence Scholarship.

\section{Limitations}

One limitation of the research in this paper is that the experiments of this paper use the pre-existing DialoGLUE few-shot splits for each dataset, following the example of prior works and to remain comparable to them (with the exception of the ablation study, which uses a separate split). However, since experiments were done only on this split, it is not necessarily the case that the results/model rankings are transferable to other splits as well (although it is worth noting from Figure \ref{fig:resampling} that performance on the random ablation split is very similar to the DialoGLUE split, and the model ranking remains the same). This limitation is not the case with GoEmotions, whose results are given as averages across three random splits. Another limitation is the relatively small number of runs/seeds (only 3) due to limitations on compute.

One further limitation is that the experiments are all performed on English-language data.

\bibliography{anthology,custom}
\bibliographystyle{acl_natbib}

\appendix

\section{GenBench Evaluation Card}

\newcommand{\tabularwidth}{\columnwidth}

\newcommand{\expone}{$\square$}
        
\renewcommand{\arraystretch}{1.1}         
\setlength{\tabcolsep}{0mm}         
\begin{tabular}{|p{\tabularwidth}<{\centering}|}         
\hline
               
\rowcolor{gray!60}               
\textbf{Motivation} \\               
\footnotesize
\begin{tabular}{p{0.25\tabularwidth}<{\centering} p{0.25\tabularwidth}<{\centering} p{0.25\tabularwidth}<{\centering} p{0.25\tabularwidth}<{\centering}}                        
\textit{Practical} & \textit{Cognitive} & \textit{Intrinsic} & \textit{Fairness}\\
\expone\hspace{0.8mm}		
& 		
& 		
& 		

\vspace{2mm} \\
\end{tabular}\\
               
\rowcolor{gray!60}               
\textbf{Generalisation type} \\               
\footnotesize
\begin{tabular}{m{0.21\tabularwidth}<{\centering} m{0.2\tabularwidth}<{\centering} m{0.13\tabularwidth}<{\centering} m{0.13\tabularwidth}<{\centering} m{0.13\tabularwidth}<{\centering} m{0.2\tabularwidth}<{\centering}}                   
\textit{Composit.} & \textit{Structural} & \textit{Cross Task} & \textit{Cross Language} & \textit{Cross Domain} & \textit{Robustness}\\
& 		
& \expone\hspace{0.8mm}		
& 		
& \expone\hspace{0.8mm}		
& 		

\vspace{2mm} \\
\end{tabular}\\
             
\rowcolor{gray!60}             
\textbf{Shift type} \\             
\footnotesize
\begin{tabular}{p{0.25\tabularwidth}<{\centering} p{0.25\tabularwidth}<{\centering} p{0.25\tabularwidth}<{\centering} p{0.25\tabularwidth}<{\centering}}                        
\textit{Covariate} & \textit{Label} & \textit{Full} & \textit{Assumed}\\  
& 		
& \expone\hspace{0.8mm}		
& 		

\vspace{2mm} \\
\end{tabular}\\
             
\rowcolor{gray!60}             
\textbf{Shift source} \\             
\footnotesize
\begin{tabular}{p{0.25\tabularwidth}<{\centering} p{0.25\tabularwidth}<{\centering} p{0.25\tabularwidth}<{\centering} p{0.25\tabularwidth}<{\centering}}                          
\textit{Naturally occuring} & \textit{Partitioned natural} & \textit{Generated shift} & \textit{Fully generated}\\
\expone\hspace{0.8mm}		
& 		
& 		
& 		

\vspace{2mm} \\
\end{tabular}\\
             
\rowcolor{gray!60}             
\textbf{Shift locus}\\             
\footnotesize
\begin{tabular}{p{0.25\tabularwidth}<{\centering} p{0.25\tabularwidth}<{\centering} p{0.25\tabularwidth}<{\centering} p{0.25\tabularwidth}<{\centering}}                         
\textit{Train--test} & \textit{Finetune train--test} & \textit{Pretrain--train} & \textit{Pretrain--test}\\
& 		
& 		
& \expone\hspace{0.8mm}		

\vspace{2mm} \\
\end{tabular}\\

\hline
\end{tabular}

\;

Taxonomy taken from \citet{hupkes2023stateoftheart}.

\section{Classical vs. Neural Retriever}

In this section we compare the SentenceTransformers neural BERT-based retriever against a classic Okapi-BM25 retriever on the HWU64 and BANKING77 datasets. The setup used is the same as in the main paper, which is 20 examples in-context with regular nearest neighbor retrieval. In Table \ref{tab:classical} we can see that the classical BM25 retriever performs measurably worse than the neural SentenceTransformer retriever, indicating that semantically-aware neural retrieval provides a significant boost in performance.

\renewcommand{\arraystretch}{1.2}

\begin{table}[th!]
  \centering
  \small
  \caption{Comparison vs. Classical (BM25) Retriever}
    \begin{tabular}{lccccc}
    \toprule
\textbf{Model}          & \multicolumn{1}{c}{\textbf{BANKING}} & \multicolumn{1}{c}{\textbf{HWU}} \\
    \cmidrule(lr){2-2} \cmidrule(lr){3-3}
          & \textbf{10-shot} & \textbf{10-shot} \\
    \midrule
    LLaMA-2-7B (mpnet) & 89.45  & 87.82 \\
    LLaMA-2-7B (BM25-Okapi) & 84.90 & 84.76 \\
    \bottomrule
    \end{tabular}%
  \label{tab:classical}%
\end{table}%

\section{Fine-tuned Retriever}

The contrastively fine-tuned retriever was trained for one epoch to avoid overfitting, using three times as many negative pairs as positive pairs (roughly 5-10 mins depending on the dataset).

\begin{table}[th!]
  \centering
  \small
  \caption{Comparison of Models with Fine-tuned Retriever (20 examples in prompt), compared against non-fine-tuned performance}
    \begin{tabular}{lcccccc}
    \toprule
\textbf{Model}          & \multicolumn{1}{c}{\textbf{BANKING}} & \multicolumn{1}{c}{\textbf{HWU}} & \multicolumn{1}{c}{\textbf{CLINC}} \\
    \cmidrule(lr){2-2} \cmidrule(lr){3-3} \cmidrule(lr){4-4}
          & \textbf{10-shot} & \textbf{10-shot} & \textbf{10-shot} \\
    \midrule
    SBERT KNN & 87.40 {\scriptsize $\pm$ 0.21}  & 83.05 {\scriptsize $\pm$ 0.47} & 91.48 {\scriptsize $\pm$ 0.13} \\
    \hspace{3mm} vs. frozen & {\scriptsize + 2.0\%} & {\scriptsize + 7.6\%} & {\scriptsize + 6.64\%}\\
    OPT 13B & 87.71 {\scriptsize $\pm$ 0.18} & 83.83 {\scriptsize $\pm$ 0.83} & 91.83 {\scriptsize $\pm$ 0.22} \\
    \hspace{3mm} vs. frozen & {\scriptsize + 2.06\%} & {\scriptsize + 0.19\%} & {\scriptsize + 2.59\%}\\
    LLaMA 7B & 87.39 {\scriptsize $\pm$ 0.081} & 87.98 {\scriptsize $\pm$ 0.75} & 94.17 {\scriptsize $\pm$ 0.32} \\
    \hspace{3mm} vs. frozen & {\scriptsize - 0.24\%} & {\scriptsize + 0.43\%} & {\scriptsize + 2.44\%}\\
    LLaMA 65B & 88.93 {\scriptsize $\pm$ 0.056} & \textbf{90.12} {\scriptsize $\pm$ 0.51} & \textbf{95.62} {\scriptsize $\pm$ 0.17} \\
    \hspace{3mm} vs. frozen & {\scriptsize - 1.79\%} & {\scriptsize + 0.062\%} & {\scriptsize + 1.16\%}\\
    \bottomrule
    \end{tabular}%
  \label{tab:perftuned}%
\end{table}%

\subsection{Discussion}

We note large improvements in the pure 1-NN mode accuracy, as expected, as we are optimizing a metric that is directly correlated with 1-NN performance. With fine-tuning, the pure 1-NN setup becomes near-competitive with ConvFit, the previous SoTA. In terms of retrieval+ICL performance, we see mixed results. In general the performance delta is quite small, suggesting that there is no significant retrieval quality bottleneck. In general, the fine-tuned CLINC retriever provides the most boost, which is also the least data-scarce scenario (it is reasonable to expect the retriever fine-tuning to be more effective with more data).

\section{Overview of Negative Results}

In this section the experiments we performed that gave negative results are enumerated. Specifically, we tried several retrieval strategies with the intention of improving performance above naive nearest-neighbor retrieval.

\begin{enumerate}
    \item We tried ``balancing'' the classes in the prompt, i.e. giving a fixed $N$ examples from each of the nearest $M$ classes, where ``nearest $M$ classes'' is defined by each class's nearest example to the input instance.
    \item With CLINC150, which has a hierarchical label structure (labels are grouped into higher-level domain categories), we tried a two-step prediction process, where the LM would first predict the domain, then the individual label. The accuracy of predicting the domain was too poor.
    \item We tried clustering the datapoints and providing $N$ examples from each of the nearest clusters, again as defined by their nearest example to the input instance.
    \item We tried a ``deduplicative'' approach to try a more diverse prompt, where an example would not be added to the prompt demonstration pool if it was too similar to an existing example in the pool.
    \item We tried doing the pure nearest example approach (what is presented in the paper), but with a restriction to a fixed $M$ number of classes represented in the prompt (i.e. as we are adding examples, if we reach a certain $M$ number of classes represented in the prompt, we stop adding examples of other classes, and just fill the prompt with examples of the first $M$ classes, in order of similarity). This was to see if the LM potentially was having issues handling examples of too many classes in the prompt.
\end{enumerate}

\end{document}